\documentclass[11pt]{article}

\usepackage[final]{acl}

\usepackage{times}
\usepackage{latexsym}

\usepackage[T1]{fontenc}

\usepackage[utf8]{inputenc}

\usepackage{microtype}

\usepackage{inconsolata}

\usepackage{graphicx}
\usepackage{enumitem}
\usepackage{amsmath}
\usepackage{bm}
\usepackage{amssymb}
\usepackage{booktabs}
\usepackage{array}
\usepackage{subcaption}

\title{Hybrid Gated Attention}

\author{
  \textbf{Zekun Zhou\textsuperscript{1,2}},
  \textbf{Ruobing Xie\textsuperscript{1}\thanks{Corresponding author.}},
  \textbf{Lanrui Wang\textsuperscript{1}},
  \textbf{Weixuan Sun\textsuperscript{1}},
\\
  \textsuperscript{1}Tencent Hunyuan,
  \textsuperscript{2}Peking University,
}

\begin{document}
\maketitle
\begin{abstract}
Gated attention is an effective approach to mitigate attention sinks and enhance the representational capacity of attention. To further extend its effectiveness-efficiency Pareto frontier, we propose a Hybrid Gated Attention (HyGA) framework that contains three types of gating strategies. Specifically, these gates leverage diverse information from multiple stages of attention, and collaboratively build element-wise/head-wise gating from multiple perspectives, capturing intra-head and cross-head information interactions.
Through our hybrid gating components, HyGA could provide multi-source modulation signals, enabling more comprehensive control over information flow and improving the representational capacity of attention.
We also introduce low-rank matrix decomposition and learnable attention sink to further enhance training efficiency and stability.
In experiments, we evaluate HyGA on widely-used benchmarks based on different backbones. The experimental results show that our HyGA comprehensively improves both training loss and various downstream performances compared with Gated attention. HyGA has also been verified to achieve the best performance at different computation costs, with comprehensive model analyses for better understanding.
The proposed HyGA sheds light on a more effective, efficient, and stable attention mechanism.
\end{abstract}

\section{Introduction}

Multi-head attention \cite{vaswani2017attention} has become a core component of LLMs for contextual information interaction. 
Consequently, the information-processing capability of attention directly affects an LLM's contextual understanding and information integration ability, and thus plays a critical role in determining the model's capacity.

Recent years, a growing body of work has sought to improve the attention mechanism. One major research direction focuses on enhancing the efficiency of attention and reducing the KV-cache overhead, leading to representative approaches such as linear attention \cite{yang2023gated,yang2025gated}, SSMs \cite{dao2024transformers}, sparse attention \cite{liu2025deepseek,xu2026deepseek,yuan2025native}, and KV-sharing or compression mechanisms such as GQA \cite{ainslie2023gqa} and MLA \cite{liu2024deepseek}. 

Another important direction aims to address the stability issues associated with attention, among which the attention-sink phenomenon has been extensively studied.
Attention sink \cite{xiao2024efficient,gu2025attention,barbero2025llms,su2026attention,sun2026spike} refers to the phenomenon in which an attention module assigns a disproportionately large amount of attention mass to a small number of tokens with limited semantic relevance, such as the BOS token. These tokens act as attention sinks by absorbing excess attention, effectively allowing certain attention heads to approximate a no-op operation. However, excessively concentrated attention may reduce the effective utilization of contextual information, which should be considered. 


\begin{figure}[t]
\centering
\includegraphics[width=0.9\columnwidth]{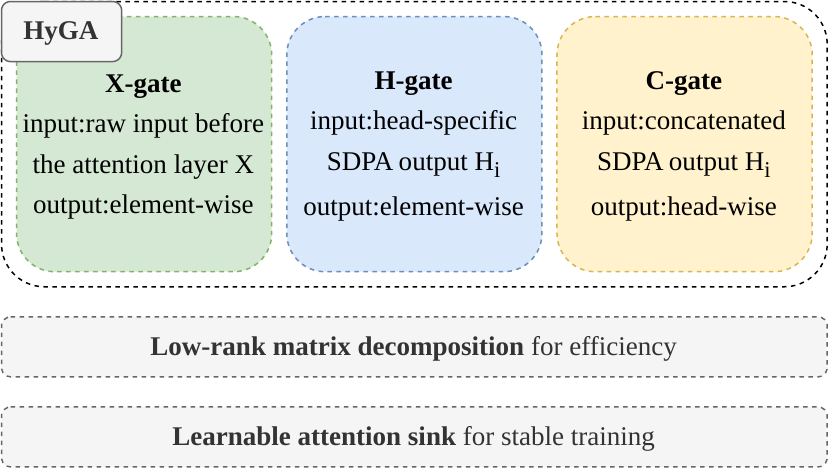}
\caption{We propose HyGA with three gates, i.e., X-gate, H-gate, and C-gate, to enhance the attention's gating strategy with more informative inputs and diverse control. It further incorporates low-rank matrix decomposition and learnable attention sinks for more efficient and stable training.}
\label{fig:Fig1}
\end{figure}

Gated attention is a representative approach for mitigating attention sinks and improving attention ability \cite{qiu2026gated}. It applies element-wise gating to Scaled Dot-Product Attention (SDPA) output, introducing non-linearity and sparsity to alleviate attention sinks and enhance model expressiveness. Nevertheless, several limitations still remain: 
(a) the original gated attention relies on the raw input X for gating, ignoring other information that is a good supplement to the current gating strategy.  
(b) Element-wise gating introduces considerable additional parameters and computational overhead, leaving substantial room for improving gating efficiency.
And (c) existing gated attention still exhibits a non-negligible BOS-token sink ratio, motivating us to further mitigate the attention-sink phenomenon for stable training.

Motivated by the above observations, we substantially extend gated attention and propose our \textbf{Hybrid Gated Attention (HyGA)}.
Specifically, our HyGA contains three well-coordinated attention gates to smartly adjust the output volume of SDPA from diverse aspects. Besides the classical attention gate controlled by the input X (\emph{\textbf{X-gate}}), we first introduce an element-wise attention gate that takes the SDPA outputs $H_i$ as the head-specific gate controller, noted as the \emph{\textbf{H-gate}}, which contains richer contextual information after attention and thereby provides additional information gains.
Since the two gates are conditioned on representations from different stages of attention, they can provide complementary modulation signals, enabling more comprehensive control over information flow and consequently improving the representational capacity of attention.
To achieve a favorable effectiveness–efficiency Pareto frontier, we apply low-rank matrix factorization to the projections of both gates. In this case, HyGA could achieve comparable training loss and downstream performance with substantially fewer attention parameters than classical Gated attention.

Besides the H-gate that focuses on the intra-head interactions to generate element-wise gating scores, we further design the cross-head gate (\emph{\textbf{C-gate}}) to capture inter-head correlations. It jointly adopts all heads' SDPA outputs to calculate head-wise gating scores, which functions as a good supplement to the above fine-grained element-wise gating.
All three gates cooperate well with each other, enriching the attention gating strategy with more comprehensive considerations on diverse features.
To prevent potential training instability and output vanishing caused by the stacked multiplication of three gates, we adopt a gate fusion strategy. We also find that HyGA could be more stable armed with learnable attention sink \cite{agarwal2025gpt}.

In experiments, we evaluate HyGA across different models on widely-used benchmarks, where the results demonstrate that HyGA consistently outperforms the original Gated attention. Through appropriate low-rank compression settings, HyGA could achieve slightly better performance with only $29\%$ of Gated attention's gating computation cost.
The contributions are summarized as follows:

\begin{itemize}[leftmargin=*]
\item We propose HyGA, which jointly adopts three hybrid gating strategies to provide both head-wise and element-wise gate scoring calculated from different factors. Equipped with learnable sink, HyGA achieves more stable training.
\item We explore different low-rank matrix factorization settings in our hybrid gates to extend the effectiveness-efficiency Pareto frontier of Gated attention methods.
\item Overall, HyGA achieves significant improvement compared to baselines with different backbones and model settings, with the low-rank technique for reducing parameters and enhancing efficiency, shedding light on more effective, efficient, and stable gated attention modules in practice.
\end{itemize}

\section{Preliminary}

In this section, we first give a brief introduction and formulation on classical attention mechanisms used in this work.

\noindent
\textbf{Grouped-Query Attention (GQA).} GQA is widely-used in popular LLMs. 
Unlike the original multi-head attention, GQA allows the queries from multiple attention heads to share a common set of key-value projections. Let $\bm{H}_i$ denote the attention output of the i-th head, the formulas are:
\begin{equation}
\bm{H}_i=
\operatorname{Softmax}\left(
\frac{\bm{Q}_i\bm{K}_{g(i)}^\top}{\sqrt{d_h}}
\right)\bm{V}_{g(i)},
\end{equation}
where $g(i)$ maps the i-th query head to its corresponding key-value group. Query heads within the same group share the same key and value representations, 
thereby substantially reducing the memory footprint of the KV cache. 

\noindent
\textbf{Multi-latent Attention (MLA).}
MLA improves attention mechanism by compressing $\bm{K}$ and $\bm{V}$ into a low-dimensional latent space. Only the low-dimensional latent representation is cached in the KV cache, and the complete $\bm{K}$ and $\bm{V}$ matrices are up-projected during attention computation. The specific formulas are as follows:
$\bm{c}_{t}^{KV}=\bm{W}^{DKV}\bm{h}_{t},
\bm{k}_{t}^{C}=\bm{W}^{UK}\bm{c}_{t}^{KV},
\bm{v}_{t}^{C}=\bm{W}^{UV}\bm{c}_{t}^{KV}$,
where $\bm{W}^{DKV}$ is the down-projection matrix and $\bm{c}_{t}^{KV}$ is the compressed latent vector shared by keys and values.
Compared with GQA, MLA achieves stronger KV cache compression. 

\noindent
\textbf{Gated Attention.}
Gated Attention \cite{qiu2026gated} refers to applying an element-wise gating operation to the output matrix after the original per-head attention. The typical gated attention function is adopted after SDPA output:
\begin{equation}
\bm{H}'_i=\bm{H}_i\odot\sigma(\bm{X}\bm{W}_i),
\;
\bm{O}=\bm{H}'\bm{W}_O.
\label{eq:ori_gated_attention}
\end{equation}
where 
$\bm{X}$ is the input matrix and $\bm{H}'$ is the gated output before $\bm{W}_O$, $\bm{W}_i$ represents the projection matrix for the i-th head, providing element-wise gating for d dimensions. It simultaneously provides nonlinear expressiveness and suppresses attention sink and massive activation, and thus could further improve the performance of both GQA and MLA.

\section{Method}
\label{sec:method}
In this work, we propose HyGA, which attempts to further optimize the effectiveness, efficiency, and training stability of gated attention. Unless otherwise specified, all equations below are presented for a single token, with the batch and sequence dimensions omitted for notational simplicity.

\begin{figure*}[t]
    \centering
    \includegraphics[width=0.82\textwidth]{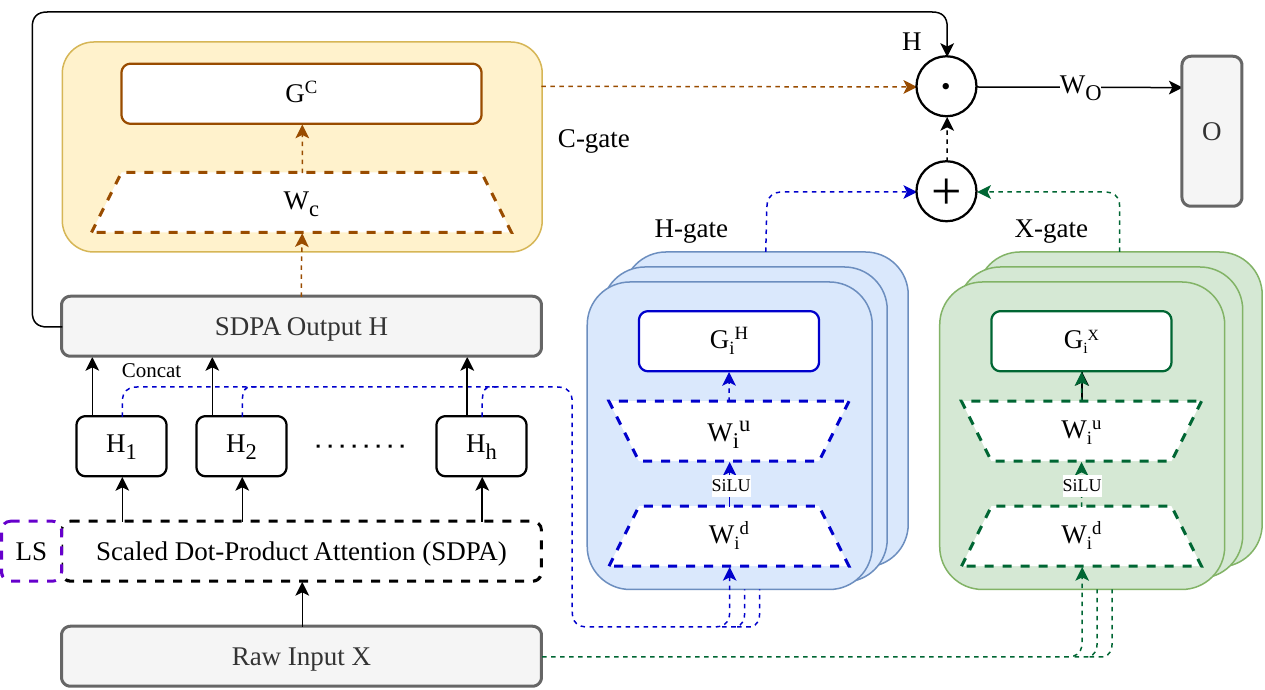}
    \caption{Overview of the proposed HyGA, which contains three gates: X-gate ($\bm{G}_X$), H-gate ($\bm{G}_H$), and C-gate ($\bm{G}_C$), armed with low-rank matrix decomposition (in H-gate and X-gate) and learnable attention sink (LS in SDPA).}
    \label{fig:main_method}
\end{figure*}

\subsection{Overall Framework}

%
As shown in Fig. \ref{fig:main_method}, HyGA mainly extends the original Gated attention with three gating mechanisms: (a) the original gate that takes the raw input $\bm{X}$ before the attention layer as input (X-gate), (b) the proposed gate that takes the output $\bm{H}$ after SDPA as input (H-gate), and (c) the proposed cross-head gate (C-gate), which captures the inter-head connections and provides head-level reweighting. We find that these three attention gates could cooperate well and jointly improve the performance.
Next, we adopt a gate fusion strategy to get our final hybrid gated attention output, avoiding excessive attenuation to ensure smooth training.
Besides, to improve training stability, we implement HyGA with learnable attention sink to provide a double guarantee for training stability.
Based on the above mechanisms, our HyGA effectively improves both model capability and training stability.

\subsection{Hybrid Gating Strategy}
\label{sec:hybrid_gating_strategy}
\subsubsection{Hybrid Gating Inputs and Functions}


The original gated attention adopts the raw input $\bm{X}$ to control the element-wise gate as stated in Eq. (\ref{eq:ori_gated_attention}).
Through an analysis of the original gated attention, we observe that the output $\bm{H}$ contains richer token-interaction information after attention, maybe providing additional information gain besides $\bm{X}$. Motivated by this observation, we propose to use the output after SDPA $\bm{H}$ as the input to control the attention gate as a supplement of $\bm{X}$. We have:
\begin{equation}
    \bm{H}_i'= \bm{H}_i\odot \sigma (\bm{X}\bm{W}_i+\text{SiLU}(\bm{H}_i\bm{W}_i^{d})\bm{W}_i^{u}).
\label{eq:baseH_attention} 
\end{equation}
For the newly added H-gate, we adopt a 2-layer MLP form with $\text{SiLU}$ \cite{elfwing2018sigmoid} to enhance nonlinear modulation. This is because the above form can more conveniently perform low-rank parameter compression by adjusting the intermediate dimension, preparing for the efficiency improvements discussed below. Here, $\bm{W}_i^{d} \in \mathbb{R}^{d \times d_{\mathrm{int}}}$ and $\bm{W}_i^{u} \in \mathbb{R}^{d_{\mathrm{int}} \times d}$ are the down-projection and up-projection gating matrices for the i-th head, respectively, where $d_{\mathrm{int}}$ denotes the intermediate dimension that controls the bottleneck of the down-then-up projection.

As can be seen from the above formulation, the original X-gate based on $\bm{X}$ and the added H-gate based on $\bm{H}$ exhibit strong functional complementarity. The former mainly captures the intrinsic features of input tokens before attention, while the latter mainly captures the features after contextual interaction through attention. 
Therefore, these two gates could well complement each other at different positions, jointly enabling more comprehensive modeling of attention outputs and finer-grained control of information flow.

\subsubsection{Fused or separate X-/H- gates.}
Different from the fused form in Eq. (\ref{eq:baseH_attention}), the hybrid X-/H- gates can also be separate:
\begin{equation}
\scalebox{0.89}{$\displaystyle
\bm{H}'_i
= \bm{H}_i \odot \sigma\bigl(\bm{X}\bm{W}_i\bigr)
\odot \sigma\Bigl(
\operatorname{SiLU}\bigl(\bm{H}_i\bm{W}_i^d\bigr)\bm{W}_i^u
\Bigr)
$}
\label{eq:baseH_attention_2}
\end{equation}
However, we choose the additive coupling form in Eq. (\ref{eq:baseH_attention}) rather than that in Eq. (\ref{eq:baseH_attention_2}). This is because the multiplicative coupling form in Eq. (\ref{eq:baseH_attention_2}) multiplies two gates together. Since the sigmoid activation function has a value range of $(0, 1)$, each gate generally acts as a suppressive modulator of the information flow. Multiplying too many gates may therefore lead to overly strong suppression, which may weaken gradient propagation and limit effective representation learning (especially with the cross-head gate in Sec. \ref{sec:cross_head_gate}).
In contrast, the fused gate in Eq. (\ref{eq:baseH_attention}) 
allows different factors to jointly control the information flow rather than to independently modulate it. This could lead to a more stable information flow and stronger capacity.
\subsubsection{Approaching the Effectiveness-Efficiency Pareto Frontier}

While pursuing improved model expressiveness, we also attempt to minimize the computation cost to extend the Pareto frontier between effectiveness and efficiency. 
Although the two gating mechanisms capture complementary aspects of the input, we hypothesize that the information required to generate their gating scores lies in a low-dimensional subspace. 
We therefore use low-rank factorization to reduce computation and parameter cost while preserving most of the gating capacity.

Specifically, we further decompose the X-based gate using the same low-rank matrix factorization adopted for our H-gate, also with a $\text{SiLU}$ activation function in the middle to provide nonlinearity.
By adjusting the intermediate dimension of $\bm{W}^{d}$/$\bm{W}^{u}$ in the two gates, we can control the maximum ranks of the two projections to balance effectiveness and efficiency. The formulation is given as follows:
\begin{equation}
\scalebox{0.75}{$\displaystyle
\bm{H}'_i
= \bm{H}_i \odot \sigma\Bigl(
\operatorname{SiLU}\bigl(
\bm{X}\bm{\bar{W}}_i^{d}
\bigr)\bm{\bar{W}}_i^{u}
+
\operatorname{SiLU}\bigl(
\bm{H}_i\bm{W}_i^{d}
\bigr)\bm{W}_i^{u}
\Bigr)
$}
\label{eq:low_rank_hybrid_gate}
\end{equation}
where $\bm{W}_i^{d}$, $\bm{W}_i^{u}$ and $\bm{\bar{W}}_i^{d}$, $\bm{\bar{W}}_i^{u}$ are different groups of weighting matrices. The maximum rank of each gating matrix can be controlled by adjusting the intermediate dimension. We find that appropriately reducing the ranks of both groups of matrices could better reduce the number of parameters and computational cost while largely preserving representational capacity, compared to the original gated attention. The resulting effectiveness-efficiency Pareto frontier is reported in the detailed experimental results in Section~\ref{sec:Pareto}.

\subsection{Cross-Head Gating Strategy}
\label{sec:cross_head_gate}


Although the H-gate and X-gate enable a more fine-grained, hybrid, and efficient attention gating, the gating strategy still depends on the intra-head element interactions for each head's output separately, while inter-heads interactions are relatively neglected.
In multi-head attention, different heads do not express information completely independently.
Instead, information from different heads can interact with each other and be expressed collaboratively.
Therefore, we hypothesize that the gating score of one head should also consider the opinions of other heads for reference.
Based on this hypothesis, we further propose the cross-head gating module (i.e., C-gate).

Specifically, we have the concatenated attention outputs of all heads $\bm{H} = \mathrm{concat}\{\bm{H}_1,\bm{H}_2,$
$..., \bm{H}_h\}$. We use $\bm{H}$ as the input and apply a matrix transformation to obtain our head-wise gating as a supplement, formulated as:
\begin{equation}
\scalebox{0.8}{$\displaystyle
\begin{aligned}
\bm{H}'_i
={}& \bm{H}_i \odot \sigma\Bigl(
  \bigl(\operatorname{SiLU}
    (\bm{X}\bm{\bar{W}}_i^{d})\bm{\bar{W}}_i^{u}
  \bigr)
  + \bigl(\operatorname{SiLU}
    (\bm{H}_i\bm{W}_i^{d})\bm{W}_i^{u}
  \bigr)
\Bigr) \\
&\odot \sigma\Bigl(
  \operatorname{Broadcast}\bigl(
    (\bm{H}\bm{W}_c)_i
  \bigr)
\Bigr).
\end{aligned}
$}
\label{eq:cross_head_gate2}
\end{equation}
Here, $\bm{W}_c \in \mathbb{R}^{hd \times h}$ denotes the transformation matrix for computing the cross-head gate for h heads based on all heads' elements. Note that $\bm{H}\bm{W}_c$ is an h-dimensional vector, and $\mathrm{Broadcast}((\bm{H}\bm{W}_c)_i)$ denotes broadcasting its i-th element to the shape of $\bm{H}_i$, indicating that this gate provides one gate score for all feature dimensions of i-th head. The reasons are: a) the C-gate is supplementary to the above element-wise X-gate and H-gate and thus should not bring in much additional computation, and b) the cross-head interactions are supposed to provide coarse-grained inter-head reweighting.

The cross-head gating strategy integrates information across heads and performs head-wise gating based on the integrated input. Since it can perceive the global head state during gating, it can dynamically adjust the relative contributions of different heads and encourage collaborative information modeling across heads. 
In conclusion, our C-gate mainly regulates coarse-grained head-wise information across heads, while the X-/H- gates mainly capture fine-grained element-wise information within each head from different aspects. Together, all hybrid gates capture distinct modulation signals and exhibit limited functional overlap. 

\subsection{HyGA with Learnable Attention Sink}
\label{sec:method_sink}


The above methods improve the expressiveness of the gated attention module through the combination of multiple mechanisms. However, beyond representational expressiveness, training stability is also a critical concern in industrial-scale training of large language models. Attention sinks \cite{xiao2024efficient} and massive activations are important factors that affect training stability. In the original work on Gated attention, the authors also mention that the head-specific gating mechanism of gated attention can introduce sparsity and mitigate attention sinks and massive activations.

However, our experimental results show that, when using gated attention alone, its mitigation of the sink ratio is still not perfect, and massive activations occasionally occur in production-scale training. Therefore, inspired by GPT-OSS \cite{agarwal2025gpt}, we implement HyGA with learnable sinks to provide an additional safeguard for training stability. We find that adding learnable sinks on HyGA can further reduce sink ratio, make hidden states significantly smoother, and effectively alleviate massive activations, even achieve slight loss advantages. 

\section{Experiments}
\label{sec:experiments}


In this section, we conduct a series of experiments to verify the effectiveness of our HyGA with three questions: (RQ1) Does HyGA outperform the original gated attention on different backbones and benchmarks (in Sec.\ref{sec:main_results})? (RQ2) Are all components of HyGA effective (in Sec.\ref{sec:ablation})? (RQ3) Can HyGA achieve the effectiveness-efficiency Pareto frontier with stable training (in Sec.\ref{sec:Pareto} and \ref{sec:stability})?

\begin{table*}[!thbp]
\centering
\begin{tabular}{p{2.4cm}<{\raggedright}p{1.46cm}<{\centering}p{1.46cm}<{\centering}p{1.47cm}<{\centering}p{1.47cm}<{\centering}p{1.46cm}<{\centering}cc}
\toprule
Model
& CEval & CMMLU & MMLU & AGIEval
& ARC & GPQA-D & GSM8K \\
\midrule
Gated attention
& 50.45 & 50.55 & 50.75 & 30.24
& 53.79 & 16.47
& 28.23  \\

HyGA
& \textbf{51.47} & \textbf{50.59} & \textbf{51.59} & \textbf{31.00}
& \textbf{57.80} & \textbf{17.79}
& \textbf{30.45} \\
\bottomrule
\end{tabular}

\begin{tabular}{p{2.4cm}<{\raggedright}ccccp{1.08cm}<{\centering}ccc}
\toprule
Model & MATH & MBPP+
& HellaSwag & PIQA & SIQA
& NQ & TriviaQA
& \textbf{Average} \\
\midrule

Gated attention
& 15.13 & 34.92
& 62.78 & 74.08 & 53.79
& 14.24 & 38.24
& 40.98 \\

HyGA
& \textbf{16.95} & \textbf{39.77}
& \textbf{62.89} & \textbf{75.23} & \textbf{54.50}
& \textbf{15.34} & \textbf{38.80}
& \textbf{42.44} \\
\bottomrule
\end{tabular}%
\caption{Results over three runs on 14 widely-used benchmarks of MoE-5B (MLA) trained on around 500B tokens.}
\label{tab:main_results_on_gated_mla_baseline}
\end{table*}


\begin{figure*}[!hbpt]
    \centering
    \includegraphics[width=0.99\textwidth]{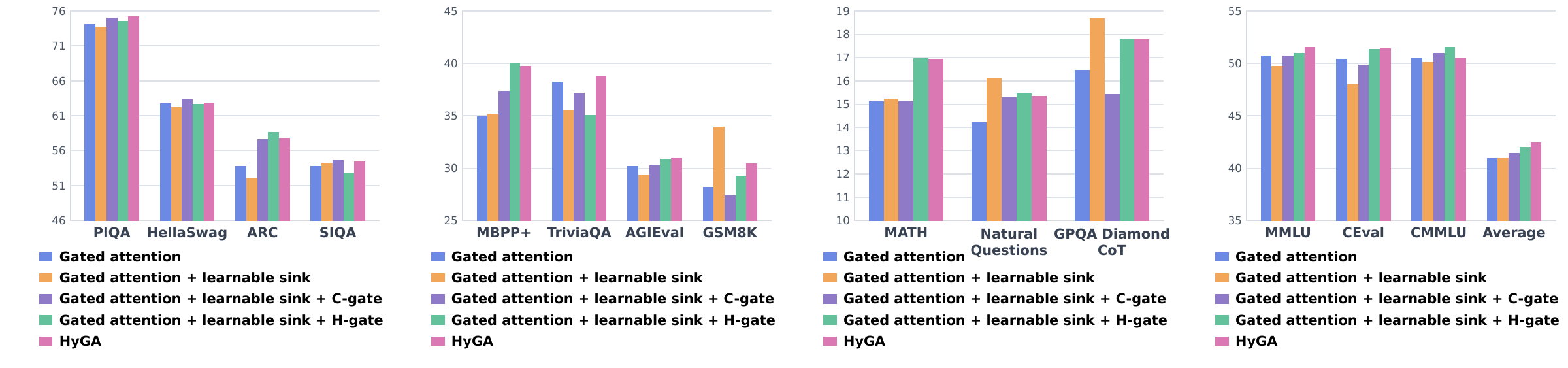}

    \caption{Ablation results for different variants of HyGA. Overall, incorporating the C-gate, H-gate, and full HyGA consistently bring benefits on 14 benchmarks (average performance gains over the baseline: $+0.49\%$, $+1.05\%$, and $+1.46\%$, respectively). These results are achieved via three runs.}
    \label{fig:ablation_results}
\end{figure*}

\subsection{Experimental Setups}

\subsubsection{Datasets.}

We evaluate models on 14 popular benchmarks, including CEval \cite{huang2023c}, CMMLU \cite{li2024cmmlu}, MMLU \cite{hendrycks2020measuring}, AGIEval \cite{zhong2024agieval}, ARC \cite{clark2018think}, GPQA-Diamond \cite{rein2023gpqa}, GSM8K \cite{cobbe2021training}, MATH \cite{hendrycks2021measuring}, MBPP+ \cite{austin2021program,liu2023your}, HellaSwag \cite{zellers2019hellaswag}, PIQA \cite{bisk2020piqa}, SIQA \cite{sap2019social}, Natural Questions \cite{kwiatkowski2019natural}, and TriviaQA \cite{joshi2017triviaqa}.
To mitigate the effects of randomness, unless otherwise specified, all benchmark scores reported below are averaged over three runs with different random seeds.

\subsubsection{Model Settings and Competitors.}

We mainly conduct experiments on an MoE \cite{jiang2024mixtral} model with nearly 1B activated parameters and 5B total parameters (noted as MoE-5B). The model adopts classical Transformer structure with MLA \cite{liu2024deepseek} used in the attention module, incorporating element-wise gated attention \cite{qiu2026gated}. It has 64 experts and 4 activated experts, armed with the Muon optimizer \cite{jordan2024muon} and trained on 500B tokens.
Besides, we also adopt another backbone of Qwen3-0.6B \cite{yang2025qwen3}, a dense model with GQA \cite{ainslie2023gqa}. It is trained with around 200B tokens via AdamW \cite{loshchilov2017decoupled}. We implement gated attention and HyGA on it for further comparisons on different LLM structures.

\subsection{Main Results (RQ1)}
\label{sec:main_results}

The experimental results of MoE-5B are shown in Table~\ref{tab:main_results_on_gated_mla_baseline}.
The results of Qwen3-0.6B are illustrated in Table \ref{tab:qwen3_six_benchmarks}. From these results, we can observe that:

(a) As evidenced by the evaluation results and loss curves of MoE-5B, our HyGA achieves consistently improvements over Gated attention across most metrics and training steps, and the overall improvement is significant (impressively $+1.46\%$ at this scale in pre-training). Note that we do not apply low-rank compression for effectiveness in the main experiments. For the loss trend, HyGA consistently maintains a lower loss throughout training, with the gap gradually widening (the loss advantage is approximately 0.012 at 60k step). These indicate the effectiveness of HyGA on MoE/MLA.

(b) For Qwen3-0.6B, its relatively small model size and trained token size make the results on some of the aforementioned fourteen benchmarks less reliable. We therefore select six relatively reliable benchmarks and report the performance gaps. We can also find that HyGA achieves the overall better performance compared to gated attention based on GQA and dense model.
Besides, after training on 200B tokens, HyGA attains a training loss approximately 0.008 lower than that of Gated GQA. Taken together, these results show that HyGA remains effective across different backbones and model settings, implying the generalization ability of HyGA.

\begin{table*}[!hbpt]
\centering
\begin{tabular}{lccccccc}
\toprule
Model & HellaSwag & ARC & PIQA & TriviaQA & NQ & MBPP+ & \textbf{Average} \\
\midrule
original GQA
& 43.93 & 25.08 & 68.39 & 13.19 & 6.93 & 7.67 & 27.53 \\

Gated attention + GQA
& 45.40 & 29.43 & \textbf{68.99} & 14.72 & 7.78 & 10.32 & 29.44 \\

HyGA + GQA
& \textbf{45.93} & \textbf{32.44} & 68.72
& \textbf{15.56} & \textbf{8.25} & \textbf{12.43}
& \textbf{30.56} \\
\bottomrule
\end{tabular}
\caption{Results over three runs on 6 widely-used benchmarks of Qwen3-0.6B (GQA) trained on 200B tokens.}
\label{tab:qwen3_six_benchmarks}
\end{table*}

\subsection{Ablation Study (RQ2)}
\label{sec:ablation}


To demonstrate that each module in HyGA makes a distinct contribution and is not fully functionally redundant with others modules, we conduct ablation studies on each components with the same settings in the main experiment (i.e., MoE-5B, MLA, 500B trained tokens). The detailed results of ablation versions on 14 benchmarks are in Figure \ref{fig:ablation_results}.

\subsubsection{Effectiveness of H-gate and Learnable Attention Sink}

We first conduct an ablation study on the H-based gating module: the gated attention baseline versus the gated attention with H-gate.
The results show that introducing H-gate yields clear improvements in most benchmarks covering different capabilities, achieving significantly better average performance. 
It implies that our H-gate can leverage the richer contextual interactions in the post-attention output $H_i$ (Section~\ref{sec:hybrid_gating_strategy}), providing additional information for element-wise modulation alongside the X-gate.
Besides, we also find that the cooperation with learnable attention sink could also bring in a slight improvement on the average score. In-depth analysis on the stability advantages of learnable sink will be discussed in Sec. \ref{sec:stability}.

\subsubsection{Effectiveness of Cross-head Gate and Gate Fusion}

%
Next, we conduct an ablation study of the C-gate by adding it to Gated attention with learnable sink. As shown in Figure~\ref{fig:ablation_results}, the C-gate yields clear overall improvements on the benchmarks.
However, we discover that multiplicatively applying three gates will excessively suppress the output activations, as stated in Sec. \ref{sec:hybrid_gating_strategy}. Therefore, we include the gate fusion strategy as the final HyGA version, achieving a more numeric-healthy gating mechanism. 
Comparing HyGA with the H-gate and C-gate variants, we observe that HyGA achieves better aggregate performance on the benchmarks than both, with a gain approximately matching the sum of their individual gains. This suggests that, under the gate-fusion strategy, the two gates may regulate information through relatively distinct pathways and jointly improve the capability of HyGA.

\subsection{Balancing Effectiveness and Efficiency of HyGA (RQ3)}
\label{sec:Pareto}


In this subsection, we discuss the balance between effectiveness and efficiency of HyGA, which could be flexibly adjusted by the low-rank matrix decomposition technique introduced in Sec. \ref{sec:hybrid_gating_strategy}. Specifically, we implement HyGA with different intermediate dimension sizes $d_{\mathrm{int}}$ of X-gate and H-gate (i.e., controlling the output dimensions of $\bm{W}^d_i,\bm{\bar{W}}^d_i \in \mathbb{R}^{d \times d_{\mathrm{int}}}$, so as to constrain the maximum rank of the projection of both gating).

Taking the H-gate 
of MoE-5B as an example, in the main experiments, we set $d_{\mathrm{int}}=d=192$ to maximize model capacity, resulting in a full-width gating structure. In this experiment, we evaluate three reduced dimensions, $d_{\mathrm{int}}\in\{16,32,64\}$, and apply the same low-rank compression scheme to the X-gate.
To assess computational efficiency and identify the effectiveness–efficiency Pareto frontier, we calculate the computation costs of attention gating modules under different $d_{\mathrm{int}}$.
Since the C-gate provides head-wise gating, it does not bring in many additional parameters (nearly 3\%). Hence, we remove the C-gate in the following experiments to more explicitly show the impact of low-rank compression on H-gate and X-gate.

\begin{figure}[!h]
\centering
\includegraphics[width=0.96\columnwidth]{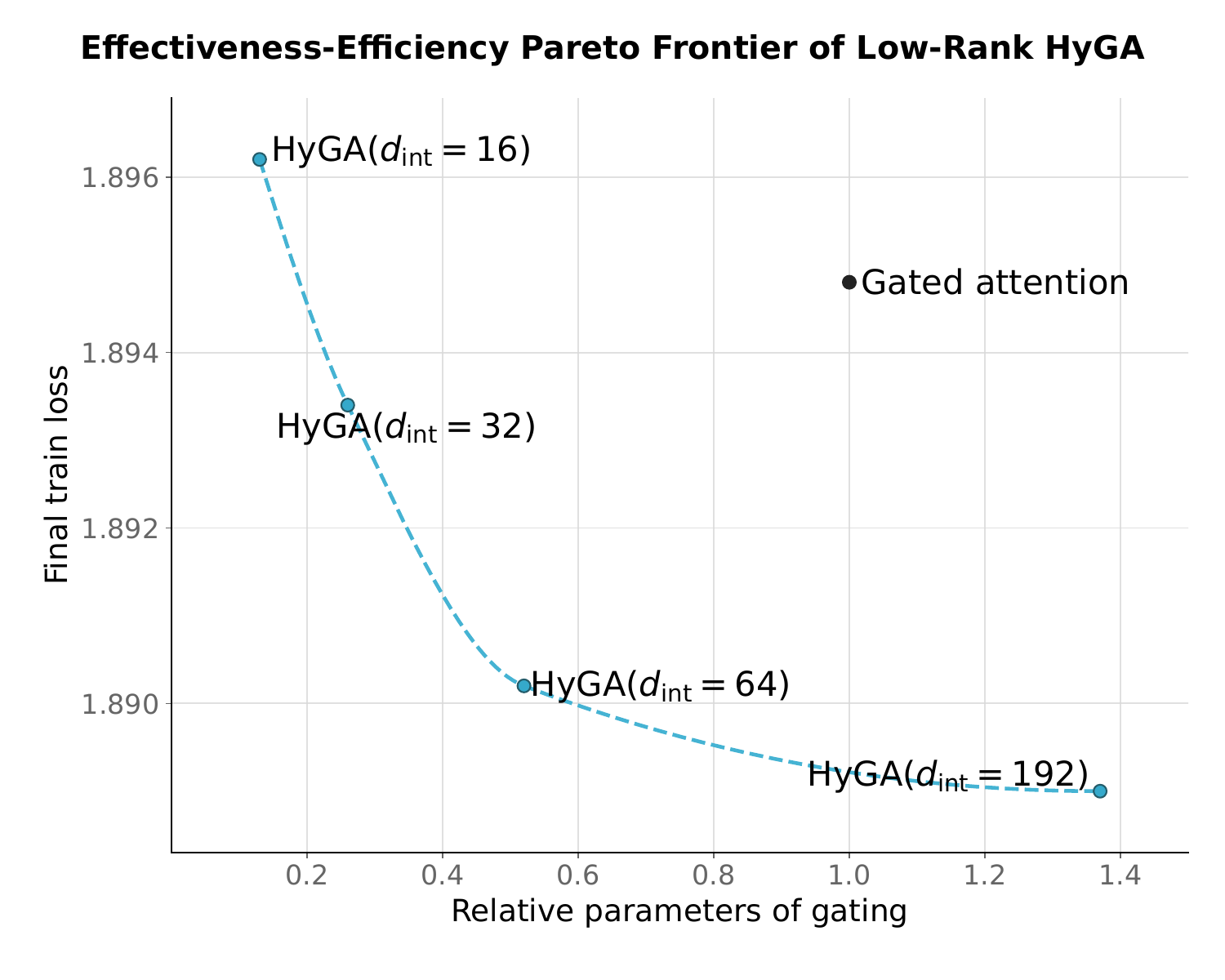}
\caption{HyGA's variants with different low-rank matrices in H-gate and X-gate could extend the effectiveness-efficiency Pareto frontier of gated attention.}
\label{fig:low_rank_pareto_frontier}
\end{figure}

We adopt the same training and evaluation settings as in the main experiments, and report the results in Fig. \ref{fig:low_rank_pareto_frontier}.

We can observe that:
HyGA achieves lower losses with larger $d_{\mathrm{int}}$ (indicating less compression rates and more computation costs).
When $d_{\mathrm{int}}=32$, HyGA (without C-gate) requires only approximately $26\%$ of the gating parameters and linear gating FLOPs of the original Gated attention. This reduces the total number of activated parameters per token by approximately $9.8\%$ and the overall model FLOPs by approximately $5.5\%$, while achieving slightly lower training loss. Moreover, we verify that HyGA with low-rank matrix factorization ($d_{\mathrm{int}}=32$) also achieves better overall performance on downstream tasks.
Moreover, we further evaluate HyGA with its parameter count and FLOPs of gating exactly matched to that of the original Gated Attention. HyGA achieves an average improvement of 1.18 points across 14 downstream benchmarks.
They demonstrate that HyGA does extend the effectiveness-efficiency Pareto frontier of the original Gated attention.
We also attempt to set different $d_{\mathrm{int}}$ for X-gate and H-gate, which seldom brings in further loss advantages.
In practice, we could flexibly set appropriate intermediate dimensions according to the practical demands.
We should highlight that in an era when many industrial LLMs push efficiency to its limits with techniques such as linear attention, \emph{reducing the model's activated parameters and FLOPs is of great importance}. HyGA offers a relatively robust and practical solution that can be adapted to most modern LLM architectures.

\subsection{Training Stability (RQ3)}
\label{sec:stability}

\begin{figure}[!t]
    \centering
    \begin{subfigure}[t]{0.95\columnwidth}
        \centering
        \includegraphics[width=\linewidth]{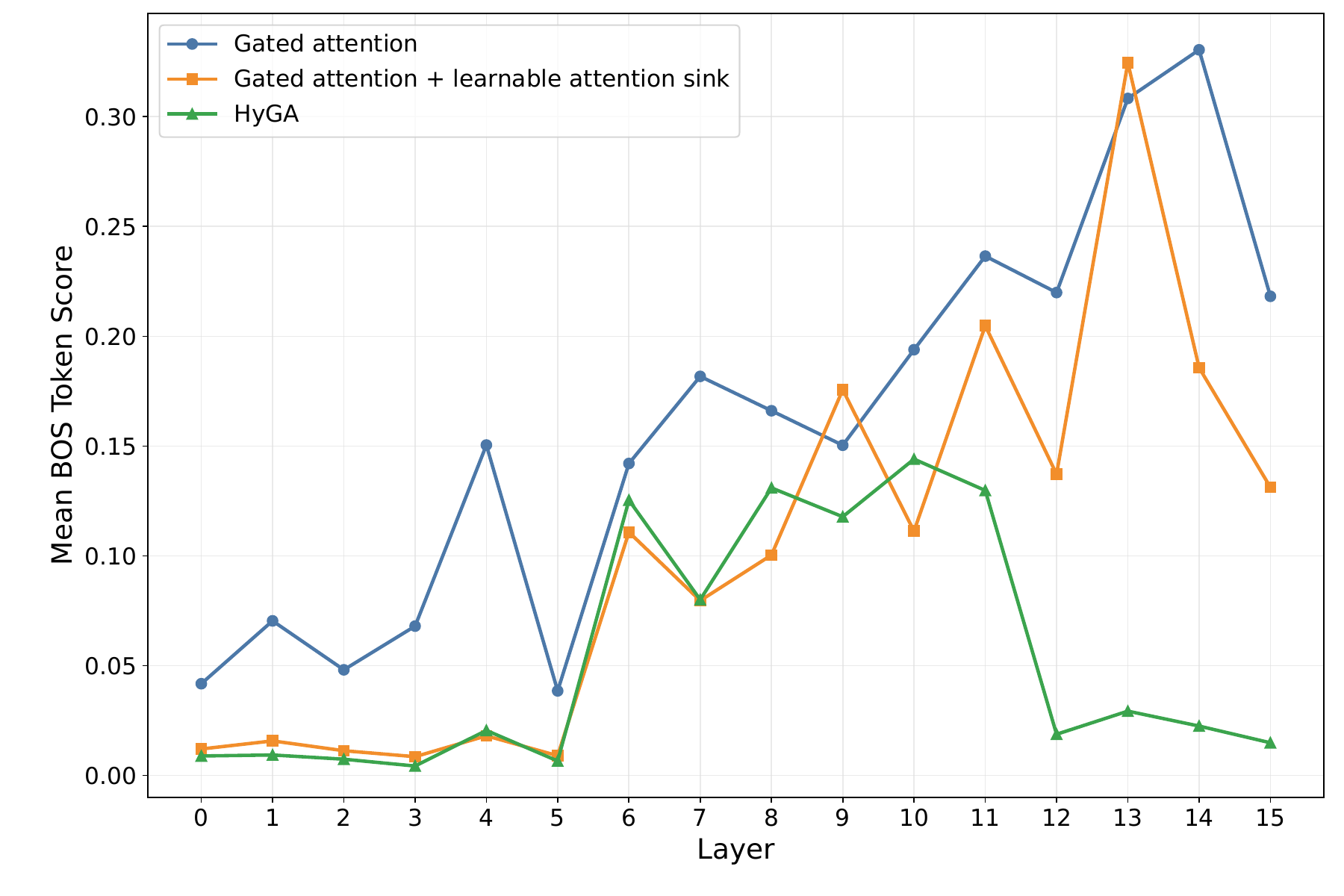}
        \caption{Layer-wise average BOS token scores.}
        \label{fig:bos_score_average}
    \end{subfigure}

    \vspace{0.5em}

    \begin{subfigure}[t]{0.95\columnwidth}
        \centering
        \includegraphics[width=\linewidth]
        {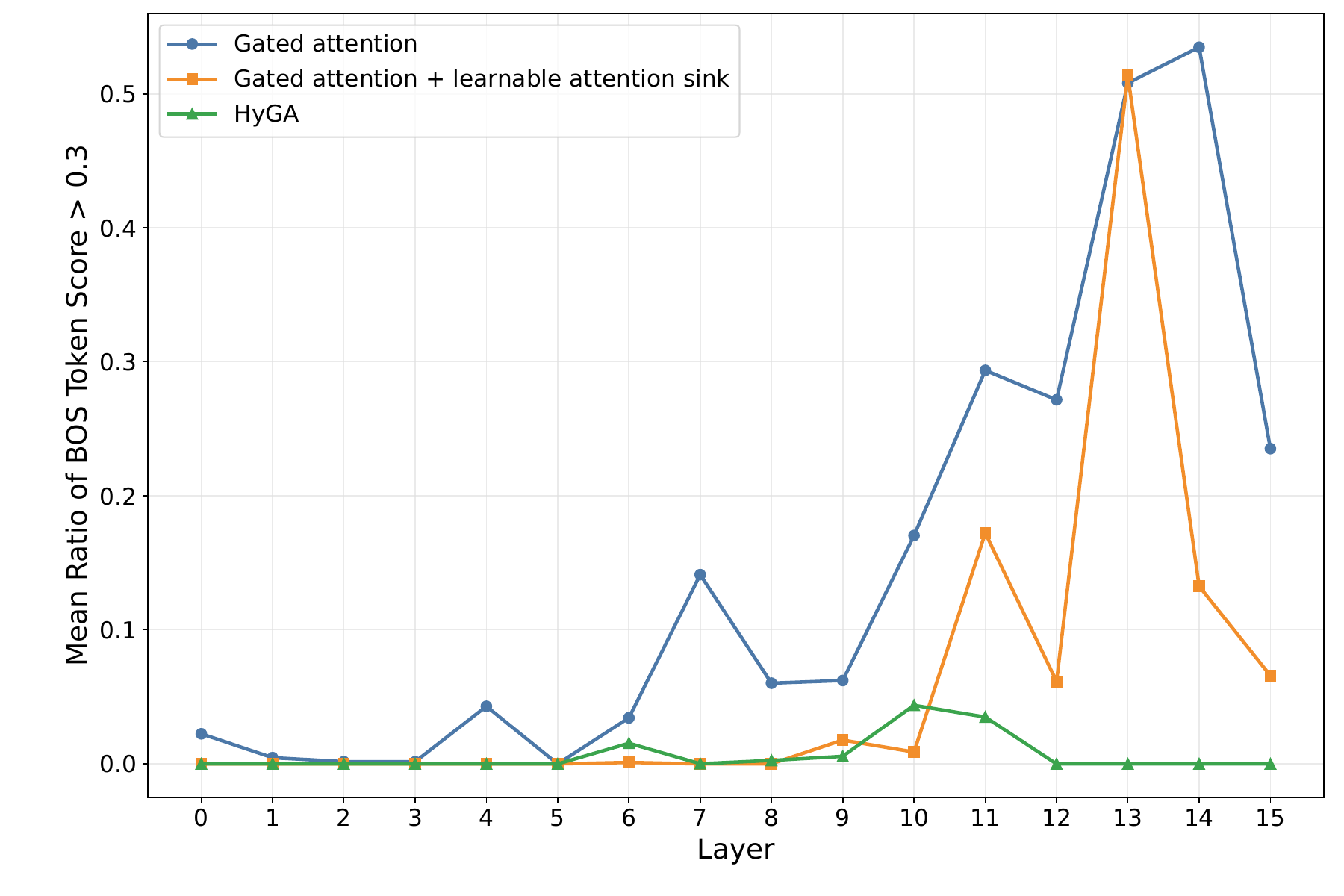}
        \caption{Layer-wise ratios of BOS token scores greater than 0.3.}
        \label{fig:bos_score_ratio}
    \end{subfigure}

    \caption{Layer-wise BOS tokens' attention score statistics of MoE-5B models:
    (a) average BOS token scores and
    (b) ratios of BOS token scores greater than 0.3.}
    \label{fig:bos_score_layer_averages}
\end{figure}

Although gated attention can largely alleviate the attention-sink phenomenon, we find that its BOS token's attention score remains relatively high compared to other tokens, indicating that attention sinks still persist. Motivated by this, we introduce learnable sinks into gated attention to further reduce sink ratios and improve training stability.
As shown in Figure~\ref{fig:bos_score_layer_averages}, HyGA with learnable sinks substantially reduces the BOS-token's attention score (especially in the last few layers). We also observe a marked reduction in massive activations, suggesting that the learnable sinks further improves training stability. Therefore, we incorporate the learnable sink into all subsequent HyGA experiments.
We further compare the BOS-token attention scores of gated attention + learnable sinks and HyGA. HyGA achieves consistently low attention sink ratios in all layers, indicating that other components of HyGA further contribute to the training stability.

\section{Related Works}

\paragraph{Gating Method in Neural Networks}
Gating is a long-standing method for controlling information flow in neural networks. Early models such as LSTMs \cite{hochreiter1997long} and GRUs\cite{cho2014learning} use gates to regulate memory updates, while later works such as GLU \cite{dauphin2017language}and SwiGLU \cite{shazeer2020glu} further introduce gating into feed-forward activations. Gating also plays an important role in modern recurrent and state-space models, including S4, Mamba, and Mamba-2 \cite{gu2021efficiently,gu2023mamba,dao2024transformers}, where it controls state updates and memory retention. 

\paragraph{Gated Attention and Its Variants}
Gated attention \cite{qiu2026gated} introduces a head-specific sigmoid gate after the SDPA output, showing that this simple modification improves performance and training stability. 
Since Gated attention, some works place gates at different locations in the attention computation, such as applying a forget gate to unnormalized attention scores \cite{lin2025forgetting} or computing gates from value states \cite{bu2025value}. Another significant direction is low-rank gate, such as gated norm \cite{qiu2026unified}. 
While these studies demonstrate the usefulness and generality of gating, they typically investigate individual design dimensions, such as gate placement or low-rank parameterization. In contrast, our work focuses on the architecture of the gate itself by jointly considering its conditioning pathways, coupling mechanism, and parameterization
, aiming to improve the model performance and training stability.

\section{Conclusion and Future Work}

In this work, we introduced HyGA with three gates depend on different inputs and granularities: X-gate, H-gate, and C-gate. Besides, we also adopt low-rank matrix decomposition and learnable sink for efficiency and training stability. Through more comprehensive information interaction capability, HyGA achieves improved performance.

In the future, we plan to scale HyGA to larger models and evaluate its effectiveness at greater scales. We will also investigate its generalizability across different attention backbones, such as linear/sparse attention architectures.

\section*{Limitations}

To assess the generalizability of HyGA across different LLM settings, we evaluate it on MoE-5B and Qwen. These models differ in scale, training-token budget, attention mechanism (MLA versus GQA), architecture (MoE versus dense), and optimizer.
Due to limited computational resources, we primarily allocate our compute budget to investigating the effectiveness and generalizability of HyGA across different model architectures. With additional resources, we plan to validate HyGA on larger models trained with substantially larger token budgets.
Besides, We have validated HyGA with both MLA and GQA. Future work will investigate whether its benefits extend to other emerging attention architectures, including linear, sparse, and hybrid attention.

\section*{Ethical Considerations}

This work focuses on architectural improvements to widely validated general-purpose large language models and does not involve human participants or the collection of personal data. Our experiments improve and train existing backbones and evaluate them on publicly available benchmarks. Therefore, the associated societal risks and ethical concerns are similar to those of existing language model research.

\bibliography{custom}

\end{document}